\documentclass[10pt,twocolumn,letterpaper]{article}

\usepackage{cvpr}      
\usepackage{algorithm}
\usepackage{algpseudocode}
\usepackage{thmtools,thm-restate}
\usepackage{amssymb} 
\usepackage{amsthm}  
\usepackage{tabularx}
\usepackage{booktabs}
\usepackage{adjustbox}

\renewcommand\qedsymbol{$\blacksquare$}

\newcommand{\ouralgorithm}{Attribute Vector Integration for Image Editing}

\newcommand{\cut}[1]{}

\definecolor{cvprblue}{rgb}{0.21,0.49,0.74}
\usepackage[pagebackref,breaklinks,colorlinks,allcolors=cvprblue]{hyperref}

\def\paperID{8799} 
\def\confName{CVPR}
\def\confYear{2025}

\title{Exploring the latent space of diffusion models directly through singular value decomposition}

\author{Li Wang\thanks{li.wang@zju.edu.cn}\\
Zhejiang University \\
\and
Boyan Gao\\
University of Oxford\\
\and
Yanran Li \\
University of Bedfordshire\\
\and
Zhao Wang
Zhejiang University\\
\and
Xiaosong Yang\\
Bournemouth University\\
\and
David A. Clifton\\
University of Oxford\\
\and
Jun Xiao \\
Zhejiang University\\
}

\begin{document}
\maketitle

\begin{abstract}
Despite the groundbreaking success of diffusion models in generating high-fidelity images, their latent space remains relatively under-explored, even though it holds significant promise for enabling versatile and interpretable image editing capabilities. The complicated denoising trajectory and high dimensionality of the latent space make it extremely challenging to interpret. Existing methods mainly explore the feature space of U-Net in Diffusion Models (DMs) instead of the latent space itself. In contrast, we directly investigate the latent space via Singular Value Decomposition (SVD) and discover three useful properties that can be used to control generation results without the requirements of data collection and maintain identity fidelity generated images. Based on these properties, we propose a novel image editing framework that is capable of learning arbitrary attributes from one pair of latent codes destined by text prompts in Stable Diffusion Models. To validate our approach, extensive experiments are conducted to demonstrate its effectiveness and flexibility in image editing. We will release our codes soon to foster further research and applications in this area.


\end{abstract}

\section{Introduction}
Diffusion models (DMs)~\cite{ho2020denoising,song2020denoising,song2020score, dhariwal2021diffusion,nichol2021improved, yang2023diffusion, croitoru2023diffusion, zhang2023text} have shown tremendous achievements in various computer vision tasks due to their ability to model complex data distributions and generate high-fidelity images. They have garnered significant attention and are extensively employed across a broad spectrum of academic and practical applications, including Text-based Image Generation~\cite{chen2024textdiffuser, xue2024raphael, zhu2023conditional, saharia2022photorealistic, ho2022cascaded}, Inverse Problems~\cite{cao2024high, li2022srdiff, amit2021segdiff, baranchuk2021label, whang2022deblurring, huang2005inverse, wunsch1996ocean}
and especially Image Editing~\cite{kawar2023imagic, ruiz2023dreambooth, bar2023multidiffusion, zhang2023sine, yang2023paint}.

Despite their advanced achievements, the latent space of the DMs has yet to be thoroughly investigated by existing researchers, although it is fundamentally crucial to image manipulation and synthesis. Straightforward editing on it usually leads to undesired results due to the complicated denoising trajectory in the diffusion inversion process. To address this problem, existing works ~\cite{dhariwal2021diffusion, ho2022classifier, kawar2023imagic, ruiz2023dreambooth, bar2023multidiffusion} tend to use classifier guidance technique or fine-tune methods (such as controlnet~\cite{zhang2023adding, zhao2024uni} and LoRAs~\cite{gandikota2023concept, hartley2024domain, pascual2024enhancing}) to manipulate the image generations.
Compared to these works on DMs, researches on GANs~\cite{radford2015unsupervised, voynov2020unsupervised,shen2020interpreting, chen2016infogan,mukherjee2019clustergan} have shown a similar manipulation on image generation but utilizing discovered semantic attribute vectors on their latent space. Thus their manipulation needs much less computational overload and performs with efficiency and flexibility.

This has sparked interest in exploring whether the latent space of DMs can be similarly harnessed. Recently, researches~\cite{kwon2022diffusion, park2023understanding, li2024self, yue2024exploring} discovered the $h$-space \cite{kwondiffusion} (i.e. feature maps of the U-Net bottleneck in DMs), has demonstrated promising semantic editing. With the help of this auxiliary space, Park \textit{et al.}~\cite{park2023understanding} utilized a pullback metric in Riemannian geometry to find semantic attribute vectors on latent space $\mathcal{X}$ to manipulate image generation, and Li \textit{et al.}~\cite{li2024self} train a learnable vector $c$ to introduce semantic attributes into generated images. However, these methods mainly focus on leveraging an auxiliary space to perform attribute manipulation rather than investigating the latent space $\mathcal{X}$ itself. As a result, they either need a data collection process or manual interpretation to identify the editing effect on limited attributes.


In light of these remarkable works, we are curious about one fundamental question: (1) Can we further explore the interpretability of latent space $\mathcal{X}$ itself and leverage it for effective and flexible image editing?

Motivated by this, we conduct a thorough investigation of the latent space $\mathcal{X}$ through a series of extensive experiments. Surprisingly, we discover that the latent space $\mathcal{X}$ in DMs shows three properties via singular value decomposition across diffusion time steps. Firstly, \textbf{small neighbourhood}, the subspaces constructed by left and right singular vectors (both are orthogonal vectors in descending order based on singular values) remain semantically similar across all time steps, which indicates arbitrary attributes destined by text prompts can be introduced into this small neighbourhood. Secondly, \textbf{attributes encoded in these singular vectors} in the form of vector values and their entangled singular values, and the residential attributes can not be changed if no new singular vectors (presenting new attributes) are added. Thirdly, \textbf{mobility in order}, assuming singular vectors always ordered along with descending of their singular values at all time steps, singular vectors accounting for attributes have mobility in orders across time steps. For example, at later time steps, those vectors responsible for coarse-grained attributes are ranked at higher places, but they will be descended to lower places at earlier time steps. For Stable Diffusion models, given a text prompt, these properties make it difficult to predict what attributes are controlled by which singular vectors at different time steps. However, they also provide an alternative way to introduce new attributes by leveraging the order mobility of singular vectors (regarded as attribute vectors in Section 3.2) between latent codes from two different time steps.    

Based on this observation, we designed a novel image editing framework to learn new attributes for Stable Diffusion Models, which are triggered by a pair of text prompts. For example, an original prompt "A photo of a male person" and a target prompt "A photo of a young male person" are paired to learn the attribute "young" in latent space. This is achieved via a careful design of singular vector integration between latent codes created by two prompts respectively at different time steps. Specifically, we design to integrate singular vectors decomposed from original latent code $x_{T_x}$ and target latent code $z_{T_x+\Delta \tau}$ in a special operation. We also propose an MLP network to predict singular values for re-weighting integrated attributes and loss terms to balance the trade-off expression of original attributes and introduced attributes. To validate our approach, we provide a \text{theoretical analysis} to demonstrate the fidelity of image editing for various attributes and undertake comprehensive experiments across diverse datasets. These results consistently illustrate the efficacy of our method, showcasing its ability to produce high-quality edits while preserving the identity fidelity of original ones.



In summary, our contributions are threefold: (1) we further explore the latent space in DMs and show three properties that they hold from the perspective of SVD on latent codes. (2) we designed a novel framework with new loss terms for efficient and flexible image editing, which performs directly on the latent space and once at a specific time step in the diffusion process. This requires no data collection process and other auxiliary spaces. (3) we provide extensive experimental validation and thorough mathematical analysis of our approach.
We believe that our new findings not only provide further insights in the latent space of DMs but also pave the way for future innovations in image manipulation. To inspire further exploration, our project will be publicly available for the research community soon. 


\section{Related Work}

\textbf{Image Manipulation in Diffusion Models.} The mainstream approaches~\cite{huang2024diffusion}, which manipulate the styles, poses or semantic contents of the generated images, are categorized as training-based methods, test-time fine-tuning methods, and training and fine-tuning free methods. A typical training-based approach~\cite{kim2022diffusionclip, wang2023stylediffusion, huang2024diffstyler} introduces a pre-trained classifier (e.g., CLIP~\cite{radford2021learning}) as guidance to adjust the gradient during the diffusion process. Another fine-tuning approach attempts to fine-tune the entire diffusion model~\cite{valevski2023unitune, choi2023custom, huang2023kv}, optimize the latent codes~\cite{mou2023dragondiffusion, shi2024dragdiffusion, nam2024contrastive, yang2023magicremover} or the text-based embedding~\cite{wu2023uncovering, dong2023prompt} to manipulate the output contents. For example, Imagic~\cite{kawar2023imagic} employs a hybrid fine-tuning method to achieve non-rigid text-based image editing by finding a representative latent for the target image. Some of the training and fine-tuning free methods~\cite{kim2023user, elarabawy2022direct, huberman2024edit, gholami2023diffusion, patashnik2023localizing, park2024shape} tried to modify the attention map or the cross-attention map to manipulate outputs. Most of these works modify the generated images in an implicit way, which is based on new training data and motivated intuitively. However, our proposed method based on latent space could be more efficient and explainable without complex extra data collection and model fine-tuning. 

\textbf{Interpretable Diffusion Models.} Although the latent space is fundamentally crucial to image manipulation and synthesis, few works have taken in-depth investigations. Some existing works~\cite{choi2021ilvr,meng2108sdedit} attempted to add explicitly guidances into latent codes to manipulate the generation results. Kwon \textit{et al.}~\cite{kwon2022diffusion} instead discovered a feature map, denoted as $h$-space, in between the bottleneck of U-Net in DMs that shows semantic correlation with text embeddings from CLIP. It can be used to learn vectors for manipulating attributes in generated images. Following this idea, Li \textit{et al.}~\cite{li2024self} designed a self-supervised approach to learn vectors in this auxiliary space for the generation of gender fairness and safe content. Park \textit{et al.}~\cite{park2023understanding} attempt to discover the local basis vectors on the latent space for editing attributes, which relies on finding the principal singular vectors from the Jacobian matrix that bridge the latent space $\mathcal{X}$ and $h$-space. Their method needs manual interpretation to discern the impact of found basis vectors. Compared to them, our exploration is steered from this auxiliary space to the latent space $\mathcal{X}$ itself, and we find properties to hold across time steps and utilize them to perform efficient and versatile editing on generated images.


\section{Method}
\begin{figure*}[h]
    \centering
    \includegraphics[width=1\textwidth]{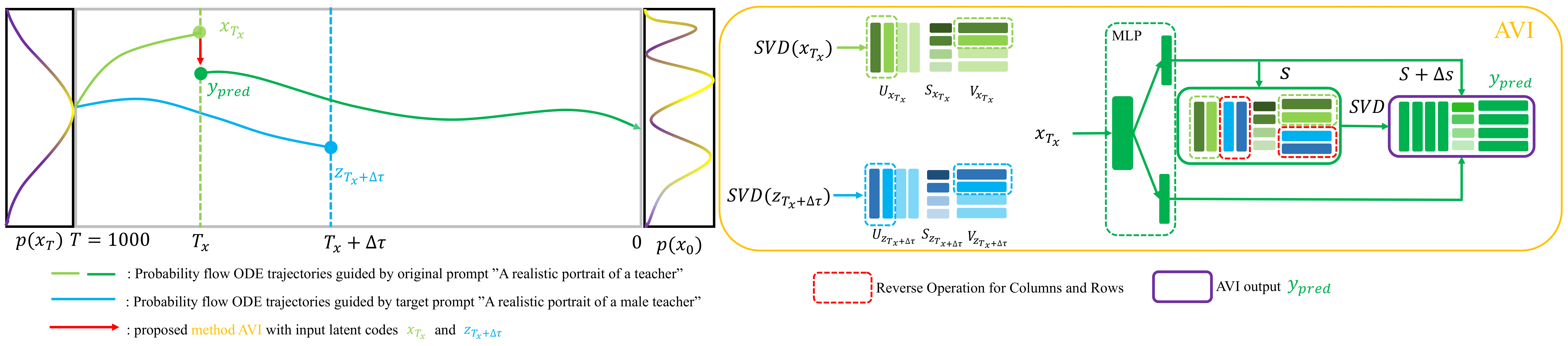}
    \caption{Our framework overview for image editing. (1) During the denoising process, we select one time step $T_{x}$ for introducing new attributes. (2) Two latent codes $x_{T_{x}}$ and $z_{T_x + \Delta \tau}$, guided by a pair of text prompts, is fed into our proposed AVI algorithm. (3) The AVI outputs a latent code $y_{pred}$ to replace $x_{T_x}$ to continue the denoising process with the guidance of the original text prompt. Note that the SVD is performed channel-wise.}
\end{figure*}

\subsection{Diffusion Models}
Diffusion models (DMs)~\cite{ho2020denoising,song2020denoising,song2020score, dhariwal2021diffusion,nichol2021improved, yang2023diffusion, croitoru2023diffusion, zhang2023text} are a class of generative models that learn to generate data by simulating a denoising process. In this process, noise is incrementally added to the data, transforming it into pure noise. The model is trained to reverse this, learning to iteratively denoise the data and generate new samples that match the original distribution. Specifically, starting from random noise $x_{T} \in N(0,I)$ the model iteratively subtracts estimated noise at each time step to obtain a denoised latent code, denoted as $x_{t-1} = x_{t} - \epsilon_{\theta}(x_t, t)$, where $\epsilon_{\theta}$ represents the U-Net model of the DMs and $\epsilon_{\theta}(x_t,t)$ is the estimated noise by the model.

The training objective is to minimize the discrepancy between the noise predicted by the model and the actual noise introduced at each time step. This optimization function can be formulated as:
\begin{equation}
\mathcal{L}_{\text{diffusion}} = \mathbb{E} \left\| \epsilon_{t} - \epsilon_\theta(x_t, t) \right\|^2 \nonumber
\end{equation}
where $\epsilon_{t} \in N(0,I)$ denotes the ground truth noise added at time step t in the forward process.

\subsection{\ouralgorithm{}}
Taking an original text prompt and a target prompt as a pair, our image editing task aims to generate a new synthesis image which similar to the image $I_x$ guided by original text prompt in terms of pose and content and contains the visual attributes provided by image $I_z$ guided by target prompt. The visual attributes could be gender, action or detail descriptions such as ``smile", ``female" and ``old". In this section, we describe the details of our Attribute Vector Integration (AVI) based Image Editing method and discuss the theoretical analysis.



\subsubsection{Attribute Vector Integration}
We discover that the representative information of attributes is captured by singular vectors decomposed from the SVD of latent codes. The generated output image could be reconstructed from an SVD formulation $\hat{y} = \hat{U} \cdot S \cdot \hat{V}$ so that our objective is to find the reasonable left and right orthogonal matrixes $U$ and $V$, the singular value vector $S$ which could fuse the visual attributes of $I_x$ and $I_z$. Motivated by this, we first sample one latent vector $x$ at time step $T_x$ and another latent vector $z$ at time step $T_x + \Delta \tau$ representing $I_x$ and $I_z$ respectively. 

The following formulation could be written:

\begin{align}
U_x S_x V_x = \text{SVD}(x),\,\,\, U_z S_z V_z = \text{SVD}(z) \nonumber
\end{align}

We construct the left attribution vector $\hat{U}$ by 
\begin{align*}
\hat{U} \gets [U_{x[:,:k]}, U^{'}_{z[:,:k]}]
\end{align*}
where $[:k,:]$ denotes selecting the top $k$ column vectors based on the magnitude of their singular values. Notably, the matrix $U^{'}_{z[:,:k]}$ is organized in a reverse column order relative to $U_{z[:,:k]}$, which is concatenated column-wise following $U_{x[:,:k]}$. Similarly, we construct the right attribute vectors but row-wise by
\begin{align*}
\hat{V} \gets  \begin{bmatrix} V_{x[:k,:]} \\ V^{'}_{z[:k,:]} \end{bmatrix}
\end{align*}
Following the construction of the attribute vectors, we apply a singular value-like set $S$ to re-weight the contributions of the attribute columns, thereby generating a latent code $\hat{y}$ by optimizing the primary objective function.

\begin{align*}
\mathcal{L}_{1}(\hat{y}, z) = \left\|\hat{y} - z \right\|^2_F, \,\,\, \hat{y} = \hat{U} \cdot S \cdot \hat{V}
\end{align*}
However, optimizing the synthetic latent code $\hat{y}$ solely toward the target latent code $z$ results in information loss from the original latent code $x$, potentially degrading output identity from $x$. To mitigate this, we enforce that the produced latent code retains an information block primarily dominated by
\begin{align*}
\mathcal{L}_{2}(\Tilde{y}, x) = \left\|\Tilde{y} - x \right\|^2_F, \,\,\, \Tilde{y} = \Tilde{U} \cdot (S + \Delta s) \cdot \Tilde{V}
\end{align*}
where $\Tilde{U}$ and $\Tilde{V}$ are sorted according to the reverse order of $\hat{U}$ and $\hat{V}$ in the column and row respectively. We will discuss the process of predicting $S$ and $\Delta s$ in the later section.
Due to the concatenation design, the issue of forgetting in editing arises as the generated images tend to lose information from the original image while also implicitly hindering optimization towards the target. This phenomenon can be understood by examining the distance between attribute vectors and the source singular value vectors. Please see the supplementary for proof of Theorem 3.1.

\begin{restatable}{theorem}{eign_distance}  \label{thm:eign_distance}
Given $x, z \in \mathcal{X}$ and their corresponding SVD, $U_x,S_x,V_x = \text{SVD}(x)$ and $U_z, S_z, V_z = \text{SVD}(z)$ where $U_x, \text{ and } U_z \in R^{M\times N}$. Let $k = \frac{N}{2}$, then the distance between attribute vectors $\hat{U}$ and its source singular value vectors $U_x$ and $U_z$ satisfy:
\begin{align*}
\left\| \hat{U} - U_x \right \|  \leq \left\| \hat{U} - U_z \right\|
\end{align*}
where $\hat{U} = [U_{x[:,:k]}, U^{'}_{z[:,:k]}]$. \\
\end{restatable}

\noindent Proof: 
\newline
\noindent Given 
\begin{align*}
\sigma^{U_x}_{max} \leq \sigma^{U_z}_{max} 
\end{align*}
where $\sigma^{U_x}_{max}$ and $\sigma^{U_z}_{max}$ denotes the maximum singular values from $U_x$ and $U_z$, respectively. The following inequality holds:
\begin{align*}
\left\|  \hat{U} - U_x \right\| \leq \left\|  \hat{U} \right\| + \left\| U_x \right\| \leq \left\|  \hat{U} \right\| + \sigma^{U_x}_{max}, 
\end{align*}
then:
\begin{align*}
\left\|  \hat{U} - U_z \right\| \leq \left\|  \hat{U} \right\| + \left\| U_z \right\| \leq \left\|  \hat{U} \right\| + \sigma^{U_z}_{max}. 
\end{align*}
\hfill\qedsymbol

Figure x shows that the maximum singular values from $U_x$ and $U_z$ increases along with the time steps, which indicates the assumption in the proof is satisfied. 

One may notice that due to this efficient attribute vector integration design, the target semantic contribution merged into the reconstructed original latent code $x$, is limited by the magnitude of the tail of singular values. The same design on the reconstructed target latent code $z$ shares this similar property. The reason behind this design is located in the attribute difference between $x$ and $z$ as they are selected from different time steps, and the order of attribute vectors floats across time steps, thanks to the mobility property. Specifically, assuming attribute vectors are always ordered along with the descending of their singular values, then attribute vectors at the higher places of the order in later time steps will be slightly descended to the lower places at earlier time steps \cite{yue2024exploring}, meanwhile the attribute vectors at the lower places of the order in later time steps will be slightly ascended to higher places at earlier time steps. Thus $x$ is selected at a later time step and the order of its current main attribute vectors will be descended in the earlier time steps, while merged attribute vectors from, $z$ picked at an earlier time step,  will be ascended, which leads to the merged attributes expressed sufficiently in results.



\subsubsection{Singular Value Prediction.} 
We assume that the necessary information for generating the synthetic image can be efficiently accessed through the singular vectors of the original and target latent codes. However, the primary challenge lies in devising a coherent approach to blend these singular vectors to capture the attribute information from both latent codes. To address this, we aim to train a neural network, $\Phi$, to produce a singular value-like matrix $S$ that blends these singular vectors, allowing the composited latent codes to inherit attributes from both the original and target ones. Directly predicting weights for this singular-vector space composition poses challenges in neural network training, as it requires balancing the trade-offs necessary for generating the desired latent code. To ease this task, we introduce an auxiliary output branch that generates a small adjustment term, $\Delta s$, which compensates for any prediction error in the singular values of $x$. This adjustment helps optimize learning while keeping the latent codes on target. Additionally, we introduce two regularizers: one to limit the distance between $S$ and $S_z$, and another between $S + \Delta s$ and $S_x$, ensuring accurate representation from both source and target perspectives. We assume that predicting singular values for $z$ involves no substantial differences:
\begin{align*}
\mathcal{L}_{3}(S, S_z)&= \left\| S - S_z \right\|^2_F \\
\mathcal{L}_{4}(S+\Delta s, S_x)&= \left\| S + \Delta s - S_x \right\|^2_F
\end{align*}
where we use square of Frobenius Norm, $ ||\cdot||_F$, to measure the distance between two matrices. By doing this we implicitly regularise the decreasing order prosperity in $S$ to align with the singular vector orders. In summary, the final objective function for learning $\Phi$ is defined as follows: 
\begin{align*}
\mathcal{L}_{AVI}(\phi) := & \mathcal{L}(\phi, \hat{y},\Tilde{y}, S, S_t, \Delta s) \label{eq:main_loss} \\ 
 = & \lambda_1 \mathcal{L}_{1}(\hat{y}, z) + \lambda_2 \mathcal{L}_{2}(\Tilde{y}, x) \\
   & + \lambda_3 \mathcal{L}_{3}(S, S_z) + \lambda_4 \mathcal{L}_{4}(S + \Delta s, S_x)
\end{align*}
where $\lambda_i$ are the hyperparameters to balance the strength of each learning term. The general idea of the training and inference is shown in Alg 4 (please see the supplementary for details) with yielding $y_{pred}$ as a result, fed into the diffusion model for further image generation. Theorem 3.1 can also be applied to the inference phase, which demonstrates output $y_{pred}$ preserves better identity fidelity from $x$ than $z$. (please see the supplementary for details). Note that due to the property of the $\Phi$ that $y_{pred}$ stays in the manifold of latent space, thus our proposed MLP network is model-agnostic.
\begin{algorithm}[t]
\caption{Attribute Vector Integration}\label{alg:eigen_man}
\begin{algorithmic}[1]
\Require Original latent code $x$, target latent code $z$, MLP network $\Phi$, top $k$, ratio $\rho$, and stage type.
\Ensure $\hat{y}, \Tilde{y}, S_x, S, \Delta s$ or $\hat{U}, \hat{V}, S, S_x, \Delta s$
\State $S, \Delta s \gets \Phi(x)$
\State $U_x, S_x, V_x = \text{SVD}(x)$ 
\State $U_z, S_z, V_z = \text{SVD}(z)$
\State $U_z^{'} \gets U_{z{[:, ::-1]}}$ \Comment{Reverse column order}
\State $V_z^{'} \gets V_{z{[::-1,:]}}$ \Comment{Reverse row order}
\State $\hat{U} \gets [U_{x[:,:k]} , (1-\rho) \cdot U_{x[:,k:]} + \rho \cdot U^{'}_{z[:,:k]}]$ \Comment{Concatenate columns}
\State $\hat{V} \gets  \begin{bmatrix} V_{x[:k,:]} \\ (1-\rho) \cdot V_{x[k:,:]} + \rho \cdot V^{'}_{z[:k,:]} \end{bmatrix}$ \Comment{Concatenate rows}
\State $\hat{y} = \hat{U} \cdot S \cdot \hat{V}$
\State $\Tilde{U} \gets \hat{U}_{[:, ::-1]}$ \Comment{Reverse column order}
\State $\Tilde{V} \gets \hat{V}_{[::-1,:]})$ \Comment{Reverse row order}
\State $\Tilde{y} = \Tilde{U} \cdot (S + \Delta s) \cdot \Tilde{V}$  
\If{stage is Training}
\State Return $\hat{y}, \Tilde{y}, S_x, S, \Delta s$
\ElsIf{ stage is Inference}
\State Return $\hat{U}, \hat{V}, S_x, S, \Delta s$
\EndIf
\end{algorithmic}
\end{algorithm}

\begin{figure*}[t]
    \centering
    \includegraphics[width=0.97\textwidth]{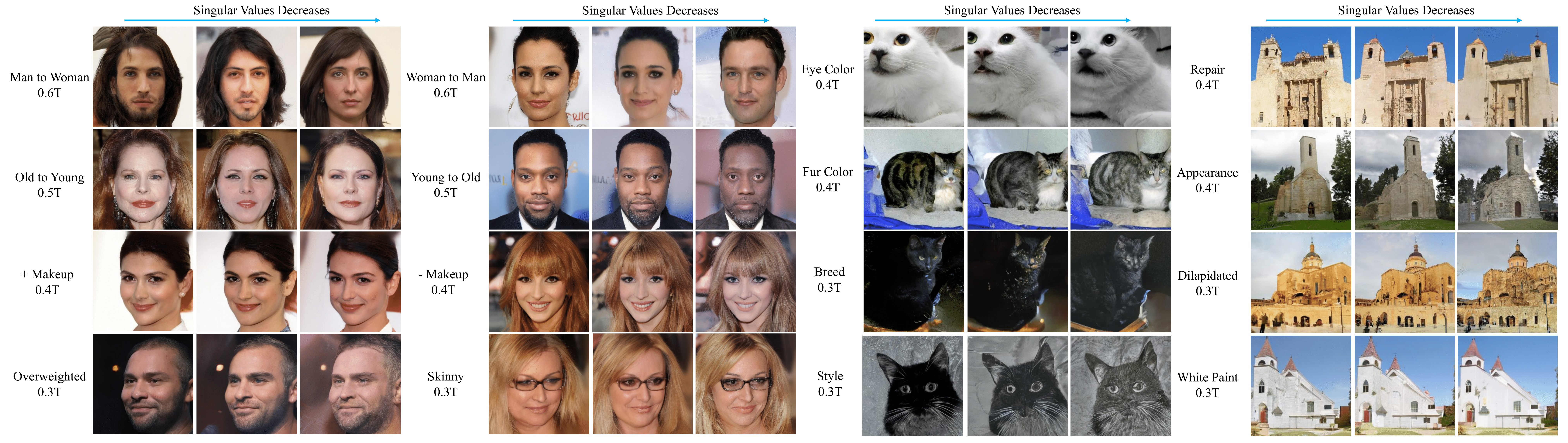}
    \caption{Impact of singular values on their main singular vectors on Unconditional Diffusion Models (CelebA-HQ dataset on the left and LSUN-Cat and LSUN-Church datasets on the right). Fine-grained attributes, such as colours and texture appearances, are changing with respect to their singular values at earlier diffusion time steps.}
\end{figure*}

\section{Findings}
In this section, we present the findings about attributes and their control capabilities within DMs. And we also dive into the properties that are held across all time steps in the denoising process.

\subsection{Attributes in Diffusion Models}

\textbf{Unconditional Diffusion Models}
In Figure 2. the left two columns present the impacts of singular values on the main singular vectors (ones with maximum singular values). We observe that at later time steps (e.g., 0.6T and 0.5T) during the denoising process (using DDPM), the slight decrease of singular values will lead to coarse-grained attributes changing, for example, gender transition and ageing. Further, the fine-grained ones, such as makeup and face shape, usually change at earlier time steps (e.g., 0.4T and 0.3T). This observation is aligned with the connections between diffusion time steps and hidden attributes \cite{yue2024exploring}. The right two columns of Figure 2 illustrate a similar observation on other unconditional diffusion models (e.g., LSUN-Cat and LSUN-Church datasets). Specifically, the fine-grained attributes, such as colours and texture appearances, are changing along with the decreasing of singular values at earlier time steps (e.g., 0.4T and 0.3T). These attributes and their effects by singular values are observed at a great chance for most samples on Unconditional Diffusion Models (e.g., CelebA-HQ, LSUN-cat, and LSUN-church datasets), as each of them shares similar structure distribution within latent space. In this work, we leave unconditional diffusion models to future exploration, but we instead delve into the conditional diffusion models (e.g., Stable Diffusion Models), as their latent spaces have a less similar structure distribution.

\textbf{Text-to-Image Diffusion Models}
We continue to investigate the impact of singular values and their singular vectors obtained from SVD of latent codes across diffusion time steps. Surprisingly, similar effects of attribute modification are observed as well. Figure 3 presents the found attributes of Text-to-Image Stable Diffusion Models (ver 2.1). The coarse-grained attributes, for example, gender, pose, background, motion, point views and appearance style, are usually observed at later diffusion time steps (e.g., 0.8T and 0.7T) during the denoising process (both DDPM and DDIM). Additionally, the fine-grained ones shown in Figure 3, such as belt textures, are observed at earlier time steps. 
Another observation is that the attributes that singular vectors are responsible for seem to be random and the residential attributes can not be changed. For example, the first row in Figure 4 shows that attributes like gender and wrinkles can be adjusted, however, this does not happen in the second row. This indicates that the adjustable attributes are related to latent codes guided by text prompts. And new attributes need to be added explicitly.


\begin{figure*}[t]
    \centering
    \includegraphics[width=1\textwidth]{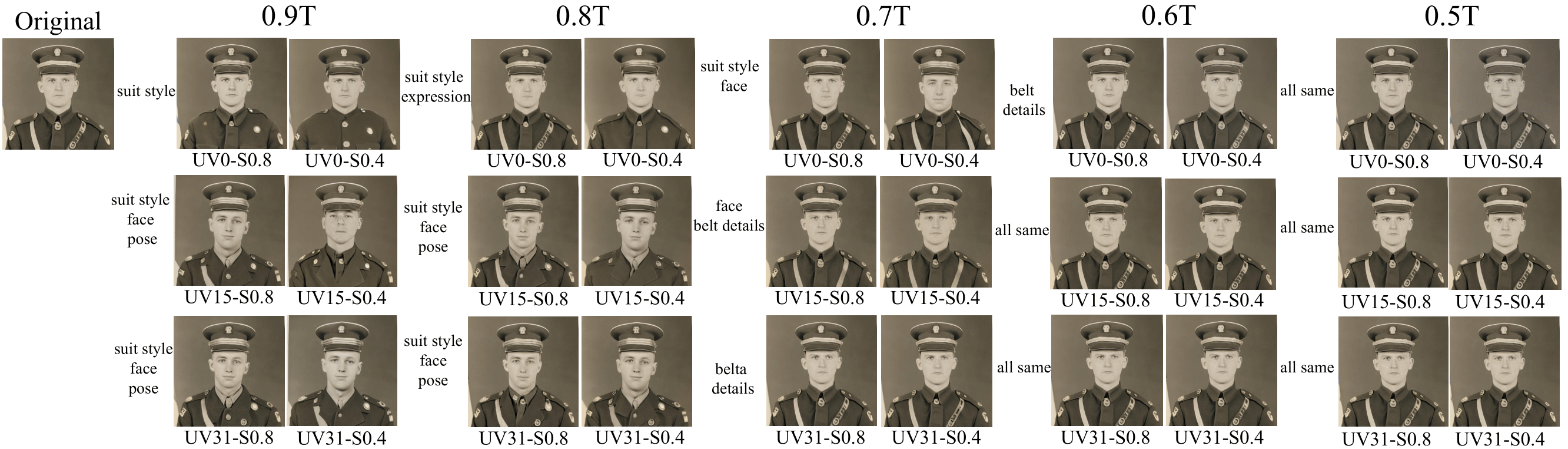}
    \caption{Representative examples shown on the attributes that one single singular vector affects across the time steps in Stable Diffusion Models (ver-2.1). Starting from the second column, the rows show the impact of singular values, and the columns present the attributes that different singular vectors may affect. Texts on the left side of images denote the attributes that a singular vector affects. It is noticeable that attribute vectors (e.g., belt details) ordered in lower places at later time steps (e.g, last row under 0.7$T$) ascended to higher places at earlier time steps (e.g, 0.6$T$)}
\end{figure*}

\subsection{Analysis of singular vectors and their singular values in Latent Space}
To further understand where the information of attributes is encoded, we conduct extensive experiments to trace them across all time steps. We discover that this information is encoded in the values of singular vectors and their magnitude of change is encoded in their paired singular values. For example, the values of singular vectors and their paired singular values slightly increase from $x_{T}$ to $x_{0}$, which also makes sense since the latent code $x_t$ gradually grows from random noise to real image (Please see supplementary for details).

In addition, we also observe that the order of singular vectors changes across diffusion time steps when assuming their orders always follow the descending of their singular values. For example, the singular vectors that affect coarse-grained attributes (e.g., identity, gender, pose, age, and clothes etc) are ordered at higher places with larger singular values at later time steps (e.g., 0.9T and 0.8T). However, at earlier time steps (e.g., 0.7T - 0.5T), those responsible for coarse-grained attributes are descended to lower places, but those for fine-grained attributes (e.g., glasses, eye pose, fatness, and colours etc.) are ascended to higher places. We refer to this as \textbf{mobility} property. This observation is also aligned that DMs generate samples in a coarse-to-fine manner \cite{pullback,yue2024exploring}. Further, we also observe that extracting the one single singular vector responsible for a semantic attribute and then using it to replace one single singular vector from another latent code will not introduce this semantic attribute. As the extracted singular vectors will be disrupted. (Please see the supplementary for details.)


Figure 4 presents the attributes that one single main singular vector that is capable of manipulating across diffusion time steps. It is noticeable that one single singular vector affects coarse-grained attributes at 0.9T time step, which may involve several fine-grained attributes. For example, the coarse-grained attribute identity is affected by multiple singular vectors (e.g., UV-15 and UV-31) which involves fine-grained attributes such as suit style, face and body pose. Across time steps, one single singular vector tends to be disentangled with other fine-grained attributes. These singular vectors with higher places in order at earlier time steps are ascending from latter places at previous time steps, which also denotes that those singular vectors of coarse-grained attributes have been descended in order. (Please see the supplementary for more details). This is the very reason that makes it possible to learn coarse-grained attributes from the latent codes with interval time steps, as latent codes at earlier time steps already contain that information in the lower places of singular vectors. The order reverse of singular vectors of latent codes at earlier time steps will put these coarse attributes to higher places again, and their magnitude will assigned with larger singular values when the predicted singular values are in descending order.

\subsection{Analysis of Geodesic Distance of Subspaces Constructed by Singular Vectors}
\textbf{The discrepancy of subspaces constructed from singular vectors maintains semantically similar in a small neighbourhood along with the generative process.} To investigate the geometry of the subspaces constructed by singular vectors, a distance metric on the Grassmannian manifold is employed to measure the distance between two subspaces. Each point on the Grassmannian manifold is a vector space, and the metric defined on it represents the distortion among various vector spaces. We employ geodesic metric \cite{choi2021not,xu2023open} to measure the discrepancy between two subspaces, which is related to the principal angles between two subspaces spanned by their columns:

\begin{align*}
d(A,B) = \sqrt{\Sigma^{p}_{k=1} \theta^{2}_{k}}
\end{align*}
where $\theta_{k}$ denotes the principal angles between two subspaces spanned by $A$ and $B$, and $p$ is the dimensionality of the subspace (in our case, $p=4$). Here the two subspaces are represented by singular vectors decomposed from the SVD of two latent codes along the generative process.

\begin{figure}[t]
    \centering
    \includegraphics[width=0.48\textwidth]{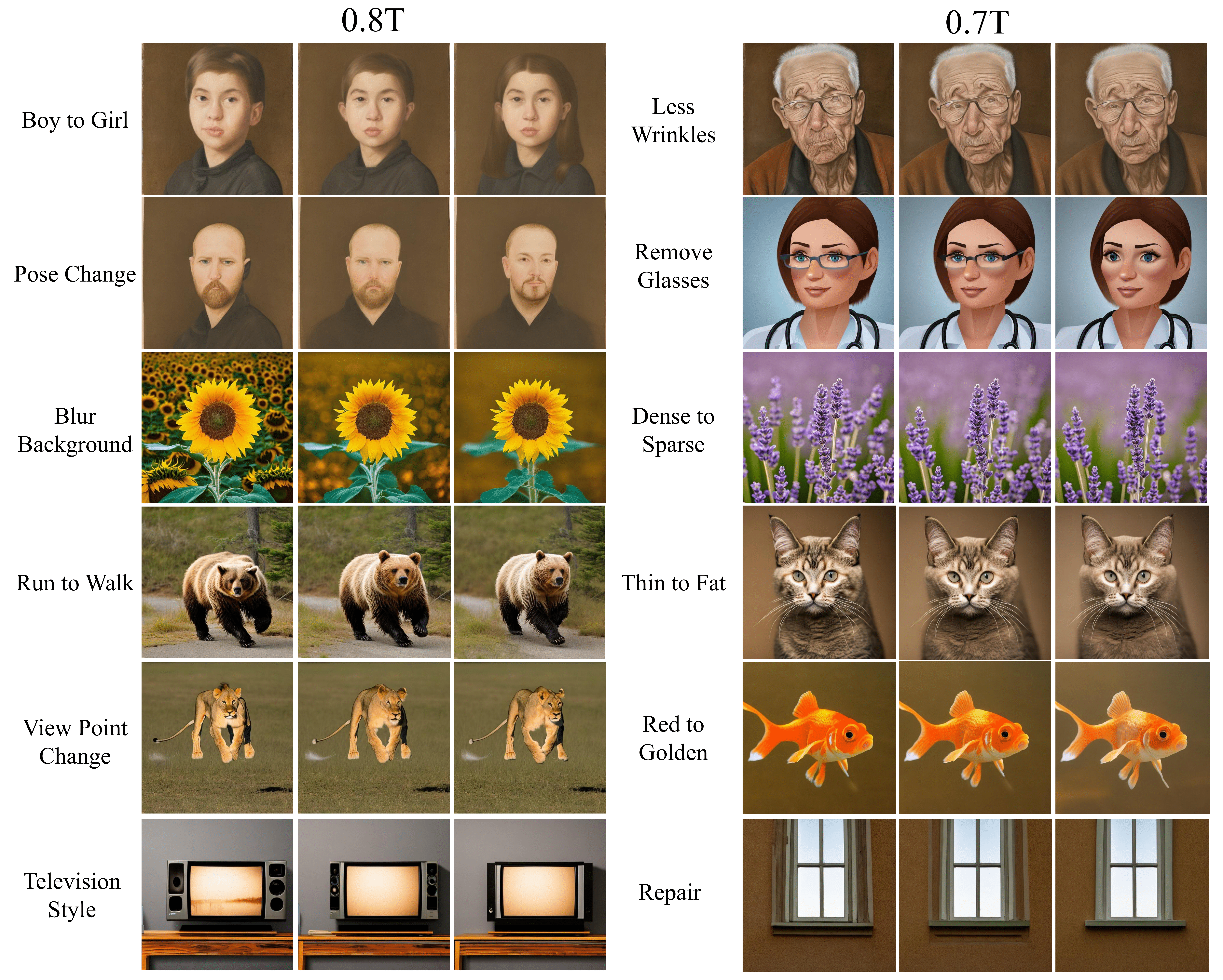}
    \caption{Impact of singular values on their main singular vectors on Stable Diffusion Models (ver 2.1).}
\end{figure}

Figure 5 demonstrates the average geodesic distance from subspaces constructed by singular vectors (30 samples) along the diffusion time steps on various datasets. It is noticeable that Unconditional Diffusion Models trained on datasets (e.g, LSUN-Cat, LSUN-Bedroom and LSUN-Church) share similar geodesic distances around 0, and Text-to-image Stable Diffusion Models (ver 2.1) with text prompt "A photo of a celebrity." and Unconditional DM on CelebA-HQ datasets share similar geodesic distance of subspace, which is around $4.3 \times 10^{-4}$. The variance of Stable Diffusion Models is less than that trained on the CelebA-HQ dataset, thus the subspaces constructed by singular vectors are in a small neighbourhood across the time steps, which also indicates that the operations on singular vectors still remain semantically similar across time steps.

\begin{figure}[t]
    \centering
    \includegraphics[width=0.5\textwidth]{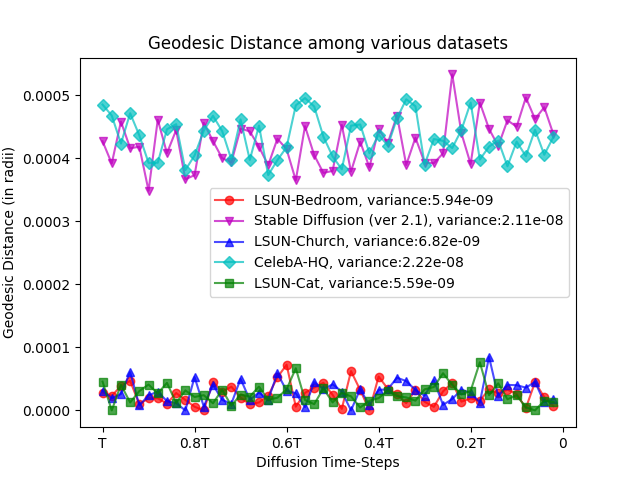}
    \caption{Geodesic Distance across subspaces constructed by singular vectors of different datasets at various diffusion time steps. It is noticeable that the variance of stable diffusion is less than diffusion models trained on the CelebA-HQ dataset, thus we tend to consider the geodesic distance on subspaces to maintain semantically similar across all time steps.}
\end{figure}
\section{Experiments on Learning Attributes}
\textbf{Experiments settings}
The proposed framework is built on Pytorch 2.1 with CUDA 12.0 in the Ubuntu 20.04 LTS system. The MLP network contains three fully connected linear layers and ReLU layers followed by two branches, in which only one linear layer is contained. The input dimension for all linear layers is fixed to 4096, and the output dimension of the last linear layer is fixed to 64, which is the same dimension as the singular value set. $k$ for the singular vector subset selection is set as 32 for our entire experiments. The hyperparameters are set as followings: for data generation, $T_{x} = 0.8T$, $\Delta \tau = -0.3T$,  $T_{z} = T_{x} + \Delta \tau$, where $T=1000$ denotes the total number of denoising time steps. the sampling number $N$ is set to 5000 when provided only one pair of latent codes, and $N$ is set to 500 when provided over 5 pairs; for training phase, batch size is set $256$, $\rho=1$, $\lambda_{1}=3$, $\lambda_{2}$, $\lambda_{3}$ and $\lambda_{4}$ are set to 10; for inference phase, $\rho$ is customized by users. The training phase uses Adam \cite{kingma2014adam} as optimizer with learning rate $1 \times 10^{-3}$. It usually reaches convergence within 5 epochs which takes around 5 mins in one single RTX 3090 Graphics Card. We set the number of steps to 50 with DDIM inversion \cite{song2020denoising} for generating real images using the Stable Diffusion Model (ver 2.1). All methods in comparison use Stable Diffusion Model version 2.1. The prompts used in all experiments follow a similar pattern that the target text prompts add or remove certain keywords from the original ones. For example, the original prompt is "A realistic portrait of a teacher", then the target prompt is "A realistic portrait of a \textcolor{red}{male} teacher". And the original prompt is "A photo of a dilapidated castle", then the target prompt is "A photo of a castle".  

\begin{table*}[t]
    \caption{Quantitative Evaluation on four attributes. The metrics under each attribute are: FID $\downarrow$, CLIP $\uparrow$ score, and LPIPS $\downarrow$.}
    \centering
    \normalsize
    \label{table1}
    \begin{adjustbox}{width=0.8\textwidth}
    \begin{tabular}{|c|c|c|c|c|c|} 
        \hline
        Method & Female & Male & Old & Young  & Runtime\\ 
        \hline
        SD~\cite{rombach2022high}   & 31.75 / 26.6  / 0.34 &  13.73 / 25.2 / 0.46 & \underline{16.47} / 24.7  / 0.39 & 35.68 / 25.9  / 0.30 & 6.2s \\  
        Park \textit{et al.} \cite{park2023understanding} & 29.27 / 25.4 / 0.07 & 16.74 / 23.2 / 0.16 & 17.06 / 23.7 / 0.18  & 37.35 / 23.9 / \underline{0.09} & 11.5s\\
        Ours & \underline{19.69} / \underline{28.4}/ \underline{0.33} & \underline{10.08} / \underline{25.4}/ \underline{0.14} & 16.93 / \underline{26.1} / \underline{0.11} & \underline{34.54} / \underline{27.59}/ 0.12 & 6.7s\\
        \hline  
    \end{tabular}
    \end{adjustbox}
\end{table*}

\textbf{Attributes Learned}
Figure 6 presents the several attributes learned via our proposed method, and it is noticeable that the structure information (e.g., identity) of the original latent code is preserved well and expected attributes are learned to alter in generated images.
\begin{figure}[h]
    \centering
    \includegraphics[width=0.48\textwidth]{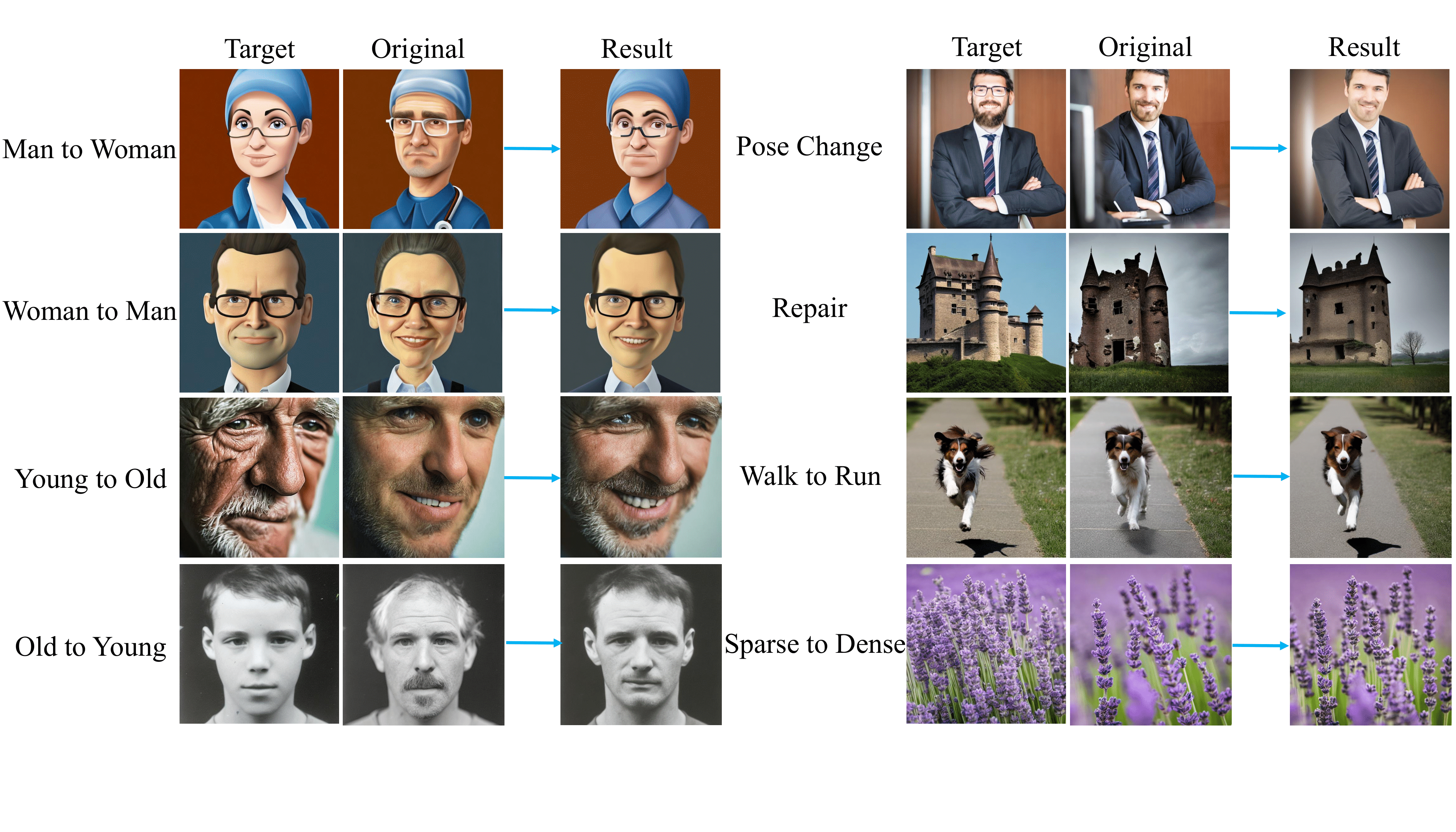}
    \caption{Examples of image edition on various learned attributes.}
\end{figure}

\textbf{Comparison to SOTA methods}
Table 1 demonstrates the performance between our method and other state-of-the-art methods ~\cite{park2023understanding,rombach2022high} on attribute manipulation. Lower FID $\downarrow$ values indicate the results present better image quality, higher CLIP $\uparrow$ scores indicate the results are more semantically aligned with the input text, and Lower LPIPS $\downarrow$ values indicate results sharing better identity fidelity of the original images. As can be seen, our method achieves the highest CLIP scores across all four attributes, lowest FID values and the lowest LPIPS values on three attributes except attribute Old and Young, respectively. Park \textit{et al.}~\cite{park2023understanding} achieves the lowest LPIPS values on attribute Young, SD~\cite{rombach2022high} achieves the lowest FID on attribute Old. 

\textbf{Interpolation on learned attributes}
Figure 7 illustrates the impact of manipulating image attributes by linearly controlling the strength of the singular vector column construction, denoted as $\rho$ in lines 6-7 of the pseudo-code of Algorithm 1. The generated image is gradually modified to the introduced attribute by adjusting the strength $\rho$, meanwhile, the identity fidelity is maintained in the transition which indicates the introduced attributes remain approximately disentangled from other semantic factors. 
\begin{figure}[h]
    \centering
    \includegraphics[width=0.45\textwidth]{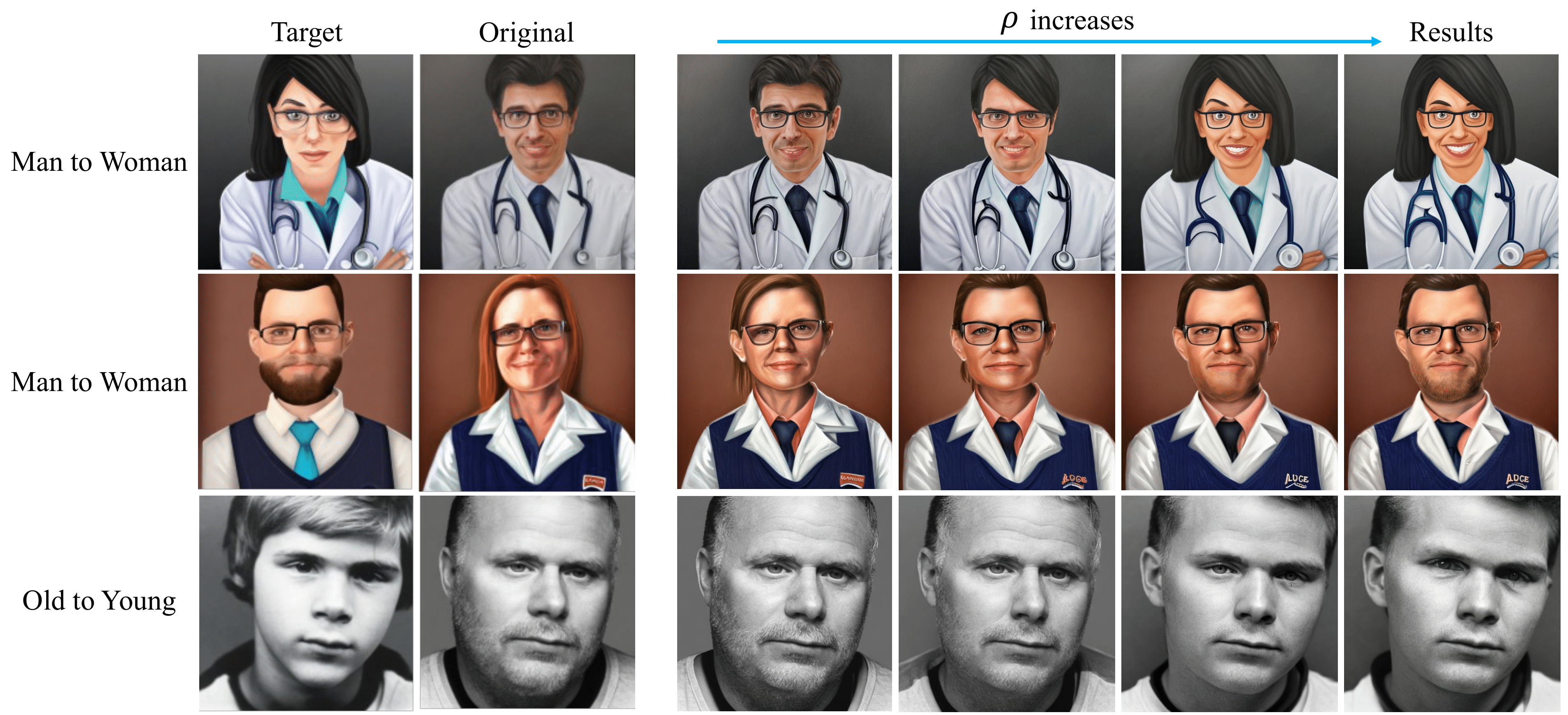}
    \caption{Linear interpolation on learned attributes.}
\end{figure}

\textbf{Ablation Study}
Figure 8 illustrates the proposed loss terms on the impact of the performance. Compared to the full model, the method without $\mathcal{L}_{1}$ leads to fewer attributes introduced into the results. As singular values $S$ are predicted to be unreasonable values for concatenated singular vectors from target latent code $z$. Thus the introduced attributes are not expressed with enough magnitudes. Similarly, the results are closer to images that target latent codes present when $\mathcal{L}_{2}$ is absent, as the original attributes are not expressed with enough magnitudes. Generated images without $\mathcal{L}_{3}$ leads to saturation loss, and the method without $\mathcal{L}_{4}$ causes image quality degraded. Due to the page limit, the discussion about hyperparameters such as loss term weights $\lambda$s, the choice of time step for original latent codes $T_{x}$ and $\Delta \tau$ for target latent codes will be shown in supplementary files. 
\begin{figure}[h]
    \centering
    \includegraphics[width=0.45\textwidth]{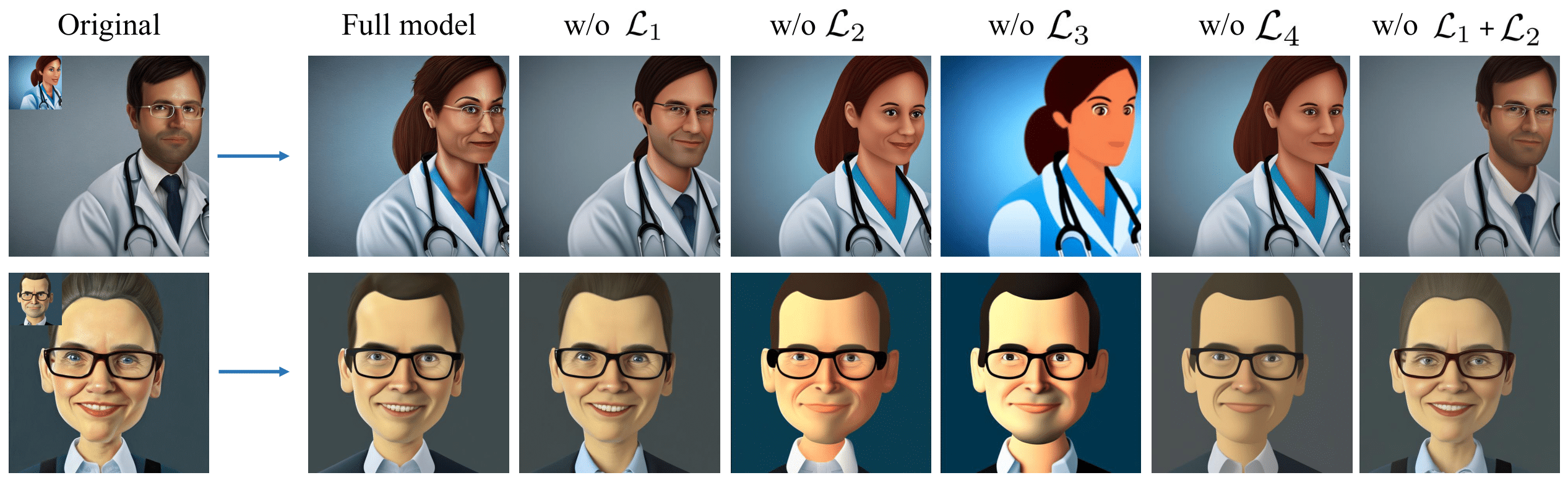}
    \caption{Ablation study on proposed loss terms.} 
\end{figure}

\section{Discussion}
In this section, we provide some intuitions and implications. It is interesting that we observe that the subspaces constructed by singular vectors remain semantically similar across diffusion time steps within DMs. The generative process seems to involve re-arranging the order of singular vectors and slightly increasing their values and singular values but not their vector directions. Thus the attributes introduced by text prompts are sort of more related to the order of singular vectors and their magnitudes. The mystery of semantic attributes could be more clarified by analysing the mechanism of order changing of these singular vectors and singular values assigned across time steps. Our method has shown the effectiveness of attributes learned from paired latent codes with time step intervals, however, it needs manual selection to find such pairs with text prompts. And it also shows sensitivity to seeds that create latent codes. In future, a promising direction could be analysing normalized singular vectors and singular values, since their directions will remain across time steps, then learning to re-order them and re-assign their singular values for attribute editing.

\section{Conclusion}
In this paper, we further explore the latent space in DMs via applying SVD directly on it and discover three properties that singular vectors hold across diffusion time steps on various datasets. Based on this finding, we propose a novel approach for image editing with attribute vector integration on latent space without data collections and any auxiliary spaces. In addition, it only performs at one specific time step and can be theoretically analysed for fidelity of image editing. Extensive experiments demonstrate the effectiveness and flexibility of our proposed image editing method. We believe our findings and fruitful insights could facilitate future research and applications of the diffusion models.

\clearpage
{
    \small
    \bibliographystyle{ieeenat_fullname}
    \bibliography{arxiv}
}


\end{document}